\documentclass[runningheads]{llncs}

\usepackage{eccv}

\usepackage{eccvabbrv}

\usepackage{graphicx}
\usepackage{booktabs}
\usepackage{multirow}
\usepackage{cite}
\usepackage{url}

\usepackage[accsupp]{axessibility}  

\usepackage[pagebackref,breaklinks,colorlinks,citecolor=eccvblue]{hyperref}
 
\usepackage{orcidlink}

\begin{document}

\title{Generative Dataset Distillation Based on Diffusion Model} 

\titlerunning{Generative Dataset Distillation Based on Diffusion Model}

\authorrunning{D.~Su, J.~Hou and G.~Li et al.}

\author{Duo Su$^\dagger$\inst{1}\orcidlink{0000-0002-9607-3639} \and Junjie Hou$^\dagger$\inst{2}\orcidlink{0009-0007-4846-4081} \and Guang Li$^\star$\inst{3}\orcidlink{0000-0003-2898-2504} \and Ren Togo\inst{3}\orcidlink{0000-0002-4474-3995} \and Rui Song\inst{4,5}\orcidlink{0000-0001-7359-1081} \and \\ Takahiro Ogawa\inst{3}\orcidlink{0000-0001-5332-8112} \and Miki Haseyama\inst{3}\orcidlink{0000-0003-1496-1761}}

\renewcommand{\thefootnote}{\fnsymbol{footnote}}
\footnotetext[4]{Equal Contribution}
\footnotetext[1]{Team Lead, Corresponding Author}

\institute{Tsinghua University \\
\email{suduo@mail.tsinghua.edu.cn} \\
\and Hong Kong University of Science and Technology \\
\email{jhouar@connect.ust.hk} \\
\and Hokkaido University \\
\email{\{guang,togo,ogawa,mhaseyama\}@lmd.ist.hokudai.ac.jp}
\and Fraunhofer IVI \\
\email{rui.song@ivi.fraunhofer.de} \\
\and Technical University of Munich \\
\email{rui.song@tum.de} \\
}

\maketitle

\begin{abstract}
This paper presents our method for the generative track of \href{https://www.dd-challenge.com}{The First Dataset Distillation Challenge at ECCV 2024}. Since the diffusion model has become the mainstay of generative models because of its high-quality generative effects, we focus on distillation methods based on the diffusion model. Considering that the track can only generate a fixed number of images in 10 minutes using a generative model for CIFAR-100 and Tiny-ImageNet datasets, we need to use a generative model that can generate images at high speed. In this study, we proposed a novel generative dataset distillation method based on Stable Diffusion. Specifically, we use the SDXL-Turbo model which can generate images at high speed and quality. Compared to other diffusion models that can only generate images per class (IPC) = 1, our method can achieve an IPC = 10 for Tiny-ImageNet and an IPC = 20 for CIFAR-100, respectively. Additionally, to generate high-quality distilled datasets for CIFAR-100 and Tiny-ImageNet, we use the class information as text prompts and post data augmentation for the SDXL-Turbo model. Experimental results show the effectiveness of the proposed method, and we achieved third place in the generative track of the ECCV 2024 DD Challenge. Codes are available at \href{https://github.com/Guang000/BANKO}{https://github.com/Guang000/BANKO}.

\keywords{Dataset Distillation \and Generative Model \and  Stable Diffusion}
\end{abstract}

\section{Background}
Deep learning has achieved remarkable success, driven by advancements in powerful computational resources. Particularly since the rise of Transformers~\cite{vaswani2017attention, liu2021swin, carion2020end, brown2020language}, the scale of models and the amount of data required for their training has increased rapidly. However, this rapid growth has led to a bottleneck in deep learning, where the ever-increasing data volume outpaces the available computational resources.
To address this issue, the development of data-efficient learning techniques~\cite{sinha2022uniform,maclaurin2015gradient} has become increasingly important. Among these, dataset distillation (DD) has emerged as a promising approach to tackle the issue of large-scale data.
Dataset distillation is a method that synthesizes a small, highly informative dataset by summarizing a large amount of real data~\cite{wang2018dataset}. Models trained on this distilled dataset can achieve generalization performance comparable to those trained on the full real dataset. This approach offers an efficient solution, particularly in managing and training large datasets~\cite{yin2024squeeze, su2024d4m, cui2023scaling}.

The significant advancements in generative models have led to the development of new approaches in dataset distillation~\cite{cazenavette2023generalizing, su2024d4m, gu2024efficient}. Generative models can encode critical information from the dataset and use it to create synthetic datasets. This approach is noteworthy due to its flexibility, allowing for various manipulations of the synthetic dataset~\cite{zhang2023dataset}, and is considered a promising direction in the evolution of dataset distillation techniques.
One critical factor affecting the test accuracy of distilled datasets is the distillation budget. This budget is defined by the concept of images per class (IPC), which constrains the size of the distilled dataset. Typically, the distilled dataset is designed to have the same number of classes as the target dataset~\cite{zhao2021datasetcondensation, nguyen2021dataset}. 
Compared with other traditional dataset distillation methods, generative dataset distillation offers superior flexibility in manipulating the IPC for distilled image generation~\cite{moser2024ld3m,su2024d4m}. This flexibility suggests that the synthetic datasets produced through this method could adapt to a broader range of tasks and conditions.

In this paper, we propose a new generative dataset distillation based on Stable Diffusion, developed for the generative track of The First Dataset Distillation Challenge at ECCV 2024. Our method leverages the SDXL-Turbo diffusion model~\cite{sauer2023adversarial}, which was selected for its superior speed and quality in image generation.
Our method significantly improves efficiency in response to the challenge’s time constraints, achieving an IPC of 10 for Tiny-ImageNet and 20 for CIFAR-100, compared to the typical IPC of 1 in other diffusion models. The method also employs class-specific text prompts and post data augmentation techniques to enhance the quality of the distilled datasets. Our experimental results demonstrate the effectiveness of this approach, which earned third place in the competition.

\section{Related Work}
\subsection{Dataset Distillation}
Dataset distillation was initially proposed by~\cite{wang2018dataset}. It aims to synthesize a smaller dataset based on the original dataset while ensuring that the results of model training on the synthetic data are similar to those trained on the original dataset~\cite{li2022awesome}. Existing dataset distillation algorithms mainly include kernel-based and matching-based methods~\cite{yu2023review, lei2023survey, sachdeva2023survey}.

Kernel-based methods, which utilize ridge regression as an optimization objective, were first introduced by~\cite{nguyen2021dataset,nguyen2021kip}. These methods employ the Neural Tangent Kernel (NTK)~\cite{jacot2018neural} to generate synthetic datasets. In contrast, RFAD~\cite{loo2022efficient} leverages the Empirical Neural Network Gaussian Process (NNGP) kernel for dataset distillation, resulting in more efficient performance in classification tasks. FRePo~\cite{zhou2022dataset} enhances this approach by replacing NTKs with network features, leading to a more effective data generation method.

Matching-based methods for generating synthetic data encompass approaches based on parameter matching, performance matching, and distribution matching.
Matching gradients by training on both synthetic and original data is a widely used and highly effective method in dataset distillation. For example, approaches like DC~\cite{zhao2021datasetcondensation}, DSA~\cite{zhao2021differentiatble}, IDC~\cite{kim2022dataset}, and DCC~\cite{lee2022dataset} employ this technique. Similar to gradient matching, methods like MTT~\cite{cazenavette2022distillation}, IADD~\cite{li2024iadd}, SelMatch~\cite{lee2024selmatch}, and ATT~\cite{liu2024att} match parameters by minimizing the loss over the training trajectory on synthetic and original data. CAFE~\cite{wang2022cafe}, IDM~\cite{zhao2023idm}, and the method in~\cite{deng2024iid} utilize feature matching, ensuring that the model produces similar deep features on both generated and original data to achieve comparable performance. Similarly, \cite{sajedi2023datadam} matches the attention between two datasets, applying attention-based methods for dataset distillation. In recent years, the development of generative models has introduced new paradigms for dataset distillation.

Due to its potential applications~\cite{geng2023survey}, especially in training large models, dataset distillation has been successfully explored across various domains, including healthcare~\cite{li2020soft,li2022compressed,li2023sharing}, fashion~\cite{cazenavette2022textures, chen2022fashion, guan2023galaxy}, and trustworthy AI~\cite{xiong2023feddm, song2023federated, zhu2023calibration, tsilivis2022robust, wang2024fed}. In computer vision, dataset distillation also provides particularly significant advantages, prompting the organization of multiple workshops and challenges on the topic, such as those at CVPR 2024~\footnote{\url{https://sites.google.com/view/dd-cvpr2024/home}} and ECCV 2024~\footnote{\url{https://www.dd-challenge.com}}.

\subsection{Diffusion Models}

Generative models have experienced rapid advancements in recent years, leading to their successful application in various industries, with prominent examples including Imagen~\cite{saharia2022photorealistic}, DALL·E 2~\cite{ramesh2022hierarchical}, Stable Diffusion~\cite{rombach2022high}, and Adobe Firefly. Compared to other generative approaches like GANs~\cite{goodfellow2014generative} or VAEs~\cite{kingma2013auto}, diffusion models often demonstrate more stable and versatile generative capabilities. These models operate by gradually adding random noise to real data and then learning to reverse this process, thereby recovering the original data distribution.

Particularly in the field of computer vision, diffusion models have been successfully applied to a wide range of tasks~\cite{yang2023diffusurvey}. For instance, SR3~\cite{saharia2022image} and CDM~\cite{ho2022cascaded} leverage diffusion models to enhance image resolution. RePaint~\cite{lugmayr2022repaint} and Palette~\cite{saharia2022palette} utilize denoising techniques for image restoration. Additionally, RVD~\cite{yang2023diffusion} employs diffusion models for future frame prediction, thereby improving video compression performance, and DDPM-CD~\cite{bandara2022ddpm} uses them for anomaly detection in remote sensing images. Unlike these applications, our approach aims to harness the power of diffusion models to achieve better performance in dataset distillation.

\subsection{Generative Dataset Distillation}

A few studies have already begun exploring the potential integration of generative models in dataset distillation, with examples such as KFS~\cite{lee2022dataset}, IT-GAN~\cite{zhao2022synthesizing} and DiM~\cite{wang2023dim,li2024local}. The inherent ability of generative models to create new data aligns well with the concept of synthesizing datasets in dataset distillation. KFS improves training efficiency by encoding features in the latent space and using multiple decoders to generate additional new samples. IT-GAN, on the other hand, generates more informative training samples through a GAN model, enabling faster and more effective model training, and thereby addressing the challenges faced in dataset distillation. DiM stores the information of target images into a generative model, enabling the synthesis of new datasets. In parallel with recent works in \cite{su2024d4m, gu2024efficient, moser2024ld3m}, our approach also integrates an effective variant of the Stable Diffusion model into the dataset distillation framework, resulting in a more efficient method for distilled image synthesis.

\section{Methods}
Our goal is to distill the dataset with limited time and resources.
We leverage the prior knowledge and controllability of the generative models to develop a dataset distillation method, which does not require any parameter optimization.
The method is capable of distilling a large number of images in a short period while maintaining high sampling fidelity. 
Theoretically, the method is general as it can distill datasets with arbitrary sizes.
The core of our approach lies in the use of SDXL-Turbo, a diffusion model trained with adversarial diffusion distillation (ADD)
strategy.
SDXL-Turbo synthesizes high-fidelity images with only 1-4 steps during the inference time.

\subsection{Overall Procedure}
\begin{figure}[t]
    \centering
    \includegraphics[width=1\linewidth]{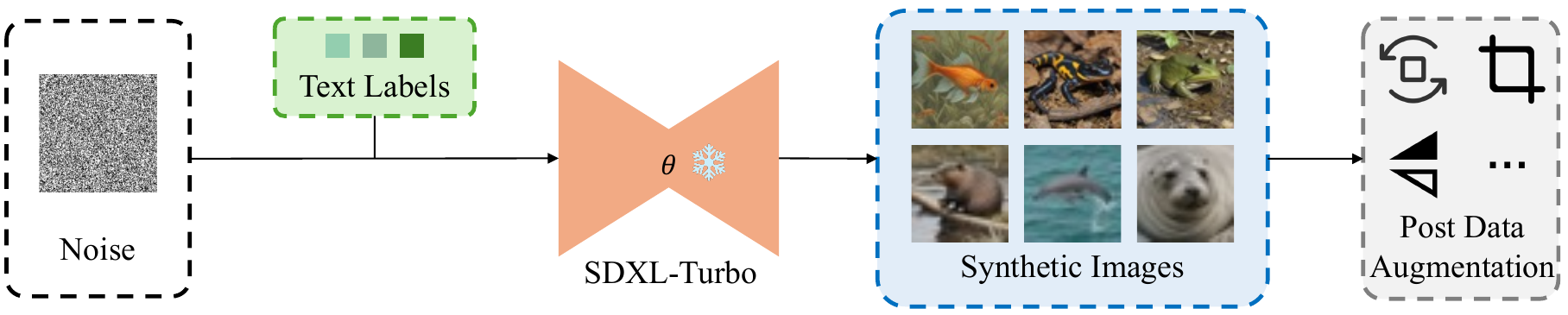}
    \caption{The pipeline of the proposed method.}
    \label{fig:overview}
\end{figure}

As illustrated in Fig. \ref{fig:overview}, we employ the Text2Image (T2I) pipeline within SDXL-Turbo, where the category labels from the original dataset are formatted as the prompts.
All of the distilled images are obtained after one-step sampling.

\subsection{Fast and High-Fidelity Dataset Distillation}
The real-time sampling of the SDXL-Turbo benefits from its ADD training strategy, which includes two objectives: 
(1) the adversarial loss involving images synthesized by the student model and the original images ($x_0$), and (2) distillation loss compares the teacher-student outputs. 
Adversarial training ensures the generated images align with the original manifolds, mitigating the blurriness and artifacts inherent in traditional distillation methods.
Distillation training effectively leverages the substantial prior knowledge embedded in the pretrained diffusion model while maintaining compositionality. 
Unlike other one-step synthesis methods (e.g., GANs, Latent Consistency Models), SDXL-Turbo retains the iterative optimization capability of diffusion models.

In adversarial training, ADD incorporates the learnable discriminator heads $\mathcal{D}_{\phi, k}$ with a pretrained Vision Transformer (ViT) feature extractor $F_k$.
For a generated sample $\hat{x}_\theta\left(x_s, s\right)$ and discriminator heads $\mathcal{D}_{\phi, k}$, the loss is determined by
\begin{equation}
\mathcal{L}_{\mathrm{adv}}^{\mathrm{G}}\left(\hat{x}_\theta\left(x_s, s\right), \phi\right)=-\mathbb{E}_{s, \epsilon, x_0}\left[\sum_k \mathcal{D}_{\phi, k}\left(F_k\left(\hat{x}_\theta\left(x_s, s\right)\right)\right)\right],
\end{equation}
whereas the timestep $s\in T_{student}=\{\tau_1,\dots,\tau_n\}$.
The discriminator is defined according to the following hinge loss:
\begin{equation}
\begin{aligned}
\mathcal{L}_{\text {adv }}^{\mathrm{D}}\left(\hat{x}_\theta\left(x_s, s\right), \phi\right)
& =\mathbb{E}_{x_0}\left[\sum_k \max \left(0,1-\mathcal{D}_{\phi, k}\left(F_k\left(x_0\right)\right)\right)+\gamma R 1(\phi)\right] \\
& \quad+\mathbb{E}_{\hat{x}_\theta}\left[\sum_k \max \left(0,1+\mathcal{D}_{\phi, k}\left(F_k\left(\hat{x}_\theta\right)\right)\right)\right],
\end{aligned}
\end{equation}
where $R1$ represents the R1 gradient penalty\cite{pmlrv80mescheder18a}.

The distillation training mainly focuses on the weighted L2 distance between the outputs of the teacher ($\hat{x}_\psi$) and student ($\hat{x}_\theta$) models.
Thus, the loss is calculated by
\begin{equation}
\mathcal{L}_{\text {distill }}\left(\hat{x}_\theta\left(x_s, s\right), \psi\right)=\mathbb{E}_{t, \epsilon^{\prime}}\left[c(t) d\left(\hat{x}_\theta, \hat{x}_\psi\left(\operatorname{sg}\left(\hat{x}_{\theta, t}\right) ; t\right)\right)\right],
\end{equation}
where sg stands for stop-gradient.
The weighting function $c(t)$ has two choices, exponential weighting or score distillation sampling (SDS) weighting\cite{poole2022dreamfusion}.

\subsection{T2I Dataset Distillation}
Considering the dataset, category labels provide crucial information that should not be overlooked. 
These labels can be embedded as conditions in diffusion models to guide the image generation process. 
During training, the discriminator of SDXL-Turbo employs a projector to extract such conditional information. 
Consequently, the trained diffusion model can leverage text labels as prompt information.

\subsection{Post Data Augmentation}
Since the Dataset Distillation challenge uses a trivial ConvNet classifier for evaluation, we aim to enhance the information richness of the distilled images without increasing computational complexity. 
Post data augmentation (PDA) is an effective approach, as it only introduces marginal overhead. 
Specifically, after synthesizing images with the diffusion model, we directly apply several augmentation methods such as image cropping, rotation, or flipping to these images. 
Please refer to Section \ref{sec:exp_detail} for detailed PDA settings.

\section{Experiments}

\subsection{Implementation Details}
\label{sec:exp_detail}
The proposed method is based on SDXL-Turbo, an effective variant of the Stable Diffusion model. 
The pretrained checkpoints come from the official Huggingface repository. 
For boosting the generation of more high-fidelity images, the model utilizes half-precision (\texttt{float16}) tensors.
The prompts consist of the corresponding category labels as mentioned.
These labels are derived from the ordered training datasets and saved in the corresponding lists. 
The hyper-parameters are set to \texttt{num\_inference\_steps=1} and \texttt{guidance\_scale=0}.
To maximize the diversity of the distilled images, Post Data Augmentation is applied after the sampling process. 
The specific parameters used for augmentations are as follows:
\begin{itemize}
    \item \texttt{RandomCrop(width=size, height=size)}
    \item \texttt{HorizontalFlip(p=0.8)}
    \item \texttt{VerticalFlip(p=0.8)}
    \item \texttt{RandomBrightnessContrast(p=0.5)}
    \item \texttt{Rotate(limit=60, p=0.8)}
    \item \texttt{RandomGamma(p=0.5)}
\end{itemize}
Considering the generation speed along with the time required for model loading and image saving, the final IPC is set to 50 for Tiny-ImageNet and 100 for CIFAR-100.
The environmental requirements of our experiments are listed as follows:
\begin{itemize}
    \item Python \(\geq 3.9\)
    \item Pytorch \(\geq 1.12.1\)
    \item Torchvision \(\geq 0.13.1\)
    \item Diffusers \(== 0.29.2\)
\end{itemize}

We follow the \href{https://github.com/DataDistillation/ECCV2024-Dataset-Distillation-Challenge}{official repository} for evaluation. 
The evaluation network, ConvNetD3-W128, will be trained for 1000 epochs.
The learning rate, momentum, and weight decay are set to 0.01, 0.9, and 0.0005 respectively.
For assessing the robustness of the method, each dataset is evaluated three times.

\subsection{Experimental Results}
\begin{figure}[t]
	\begin{center}
	\includegraphics[width=\columnwidth]{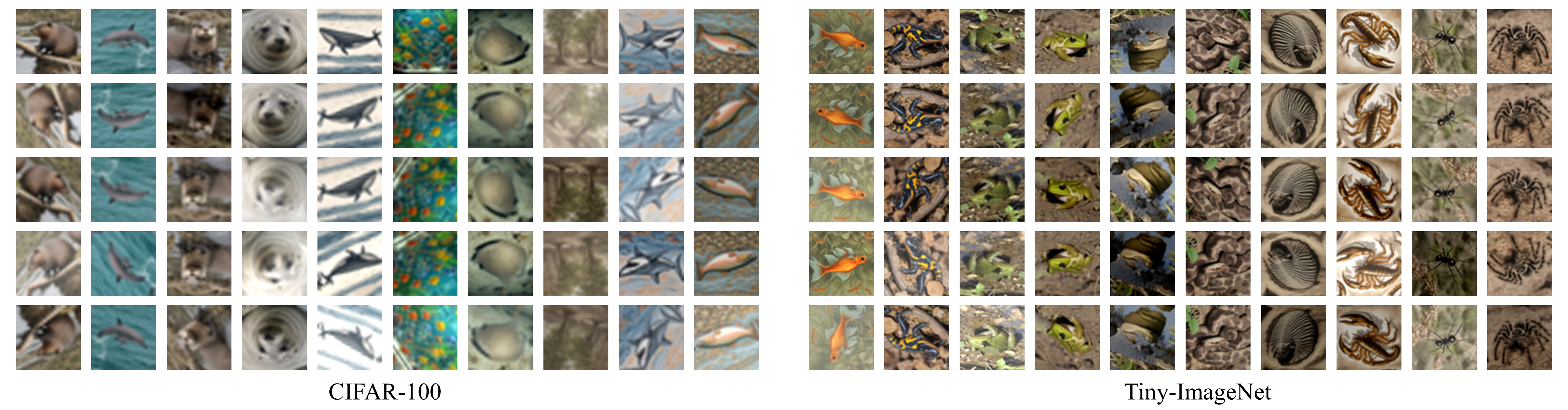}
	\end{center}
	\caption{The generated images of the top-10 classes in Tiny-ImageNet and CIFAR-100. The first row displays the images produced by the diffusion model, and the subsequent four rows represent the PDA images.
}
  \label{fig:vis}
\end{figure}

According to Table \ref{tab:abstudyss}, PDA directly increases the number of IPC and significantly improves the distilled performance.
The Accuracy increased by 0.0041 on Tiny-ImageNet, from 0.0437 to 0.0478, while in CIFAR-100, it increased by 0.0040.
The top-10 category visualization results are displayed in Fig. \ref{fig:vis}.
The first row features images generated by the diffusion model, noted for their high fidelity and realism
The remaining four rows illustrate the results of applying data augmentation to these images, which increases their diversity and improves the quality of the distilled dataset.
\begin{table}[t]
    \centering
    \caption{The model performance on the validation set with different components.}
    \begin{tabular}{lcccc}
        \toprule
        Model & \multicolumn{2}{c}{Tiny-ImageNet} & \multicolumn{2}{c}{CIFAR-100} \\
        \cmidrule(lr){2-3} \cmidrule(lr){4-5}
              & IPC & Accuracy & IPC & Accuracy \\
        \midrule
        SDXL-Turbo & 10 & 0.0437±0.0012 & 20 & 0.0097±0.0010 \\
        SDXL-Turbo + PDA & 50 & 0.0478±0.0018 & 100 & 0.0137±0.0005 \\
        \bottomrule
    \end{tabular}
    \label{tab:abstudyss}
\end{table}

\section{Discussion}

\subsection{Distribution Discrepancy}

The matching strategies in traditional dataset distillation methods ensure that the distribution of the distilled dataset aligns with the original dataset. 
However, when it comes to the generative models, significant discrepancies may arise between their distributions.
Some recent works indicate that images generated by the diffusion model perform well on the large-scale ImageNet-1K. 
This success may be attributed to two factors: (1) the images closely resemble the real-world scenes and
(2) the resolution of ImageNet-1K is similar to the generative model. 
In other words, ImageNet-1K has a resolution of 224 $\times$ 224 and the diffusion model generates images at 256 $\times$ 256.

When the dataset is set to Tiny-ImageNet, the effectiveness of dataset distillation begins to decline. 
Tiny-ImageNet is recognized as a subset of ImageNet-1K, thus, their data distributions are similar. 
However, the resolution decreases to 64 $\times$ 64, resulting in significant information loss.
When the dataset is further changed to CIFAR-100, the performance of generative model dataset distillation becomes unsatisfactory. 
This is due to a substantial distribution discrepancy between CIFAR-100 and the generated models, as well as a considerable gap in resolution. 
The Common methods, such as fine-tuning or modifying text prompts, have not effectively addressed this issue.

\section{Conclusion}
We have proposed a novel generative dataset distillation method based on diffusion model. Specifically, we use the SDXL-Turbo model which can generate images at high speed and quality. As a result, our method can generate distilled datasets with large IPC. Furthermore, to obtain high-quality distilled CIFAR-100 and Tiny-ImageNet datasets, we utilize the class information as text prompts and post-augmentation for the proposed method. The top performance during the challenge shows the superiority of our method.
In the future, we want to develop more effective distillation techniques with generative models across different datasets.

\section*{Acknowledgements}
This study was supported by JSPS KAKENHI Grant Numbers JP23K21676, JP23K11141, and JP24K23849.

\bibliographystyle{splncs04}
\bibliography{main}
\end{document}